\documentclass[conference]{IEEEtran}
\IEEEoverridecommandlockouts
\usepackage{booktabs}
\usepackage{cite}
\usepackage{amsmath,amssymb,amsfonts}
\usepackage{algorithmic}
\usepackage{graphicx}
\usepackage{textcomp}
\usepackage{multirow}
\usepackage[table,xcdraw]{xcolor}
\usepackage{mathrsfs}
\usepackage{enumitem}
\usepackage{fancyhdr}
\def\BibTeX{{\rm B\kern-.05em{\sc i\kern-.025em b}\kern-.08em
    T\kern-.1667em\lower.7ex\hbox{E}\kern-.125emX}}
\begin{document}

\title{Generalized Mixture Model for Extreme Events Forecasting in Time Series Data}

\author{\IEEEauthorblockN{Jincheng Wang}
\IEEEauthorblockA{\textit{Department of Automation} \\
\textit{Shanghai Jiao Tong University}\\
Shanghai, China \\
jc.wang@sjtu.edu.cn}
\and
\IEEEauthorblockN{Yue Gao*}
\IEEEauthorblockA{\textit{MoE Key Lab of Artificial Intelligence, AI Institute} \\
\textit{Shanghai Jiao Tong University}\\
Shanghai, China \\
yuegao@sjtu.edu.cn}
\thanks{*Corresponding author.}
}

\maketitle

\begin{abstract}

Time Series Forecasting (TSF) is a widely researched topic with broad applications in weather forecasting, traffic control, and stock price prediction. Extreme values in time series often significantly impact human and natural systems, but predicting them is challenging due to their rare occurrence. Statistical methods based on Extreme Value Theory (EVT) provide a systematic approach to modeling the distribution of extremes, particularly the Generalized Pareto (GP) distribution for modeling the distribution of exceedances beyond a threshold. To overcome the subpar performance of deep learning in dealing with heavy-tailed data, we propose a novel framework to enhance the focus on extreme events. Specifically, we propose a Deep Extreme Mixture Model with Autoencoder (DEMMA) for time series prediction. The model comprises two main modules: 1) a generalized mixture distribution based on the Hurdle model and a reparameterized GP distribution form independent of the extreme threshold, 2) an Autoencoder-based LSTM feature extractor and a quantile prediction module with a temporal attention mechanism.  We demonstrate the effectiveness of our approach on multiple real-world rainfall datasets.
\end{abstract}

\begin{IEEEkeywords}
mixture models, extreme events, generalized Pareto distribution, temporal attention mechanisms
\end{IEEEkeywords}
\section{Introduction}
Time series forecasting has been a subject of widespread research across various disciplines over the past several decades. The technique has been applied to address a broad range of real-world challenges, including climate prediction \cite{a35}, stock market analysis \cite{g17}, and traffic prediction \cite{g19}. Traditional methods like Exponential Smoothing (ES) and ARIMA were introduced to address time series forecasting challenges. In recent years, Deep Neural Networks (DNNs), particularly Recurrent neural networks(RNNs), have surpassed these traditional methods, primarily due to their ability to model time patterns deeply and non-linearly.

However, few studies have provided detailed and in-depth investigations of extreme events in time series. Extreme events such as floods, heatwaves, and droughts are infrequent, but their impact is disproportionately large, leading to catastrophic damage to properties and lives. The infrequency of such events leads to highly skewed data inputs, making their prediction in time series a formidable challenge. Traditional time series forecasting methods struggle with extreme values in predictions. And conventional deep learning algorithms typically optimize for global metrics like Root Mean Square Error (RMSE). This approach commonly results in models emphasizing predictions of the target variable's conditional mean while sidelining its extreme values for the sake of overall performance\cite{bishop2006pattern}.

In this paper, we propose a novel framework that integrates Deep Learning methods with Extreme Value Theory (EVT) for extreme value prediction. We conducted modeling and research on rainfall datasets. Motivated by the zero-inflation and heavy-tailed phenomena \cite{beyond_point_prediction} in the rainfall datasets, we utilize a mixture distribution for data modeling. Specifically, many precipitation measurements were exactly zero, while some were moderate (nonzero and non-extreme), and a few were classified as extreme. We divided the rainfall data into three parts. For extreme values, we use a reparameterization method for the Generalized Pareto (GP) distribution, which transforms it into a threshold-invariant form. For other parts, a Hurdle model and log-normal distribution are utilized. These elements are merged into a mixture model, leveraging its cumulative distribution function (CDF) for neural network prediction.

The main contributions of this article are as follows:

\begin{itemize}
    \item We employ a reparameterization method for the GP distribution, and we propose a novel deep learning framework integrating a generalized mixture model for extreme events prediction in time series data.
    \item We propose DEMMA, a network that includes an LSTM-based autoencoder for effective temporal feature extraction and a forecaster module that applies temporal attention mechanisms to predict the quantile value. A combined loss function is used to ensure both proper feature extraction and the capturing of tail distribution.
    \item Experiments on rainfall datasets show our framework's superiority over other methods in extreme value prediction and overall forecasting. 
\end{itemize}

To the best of our knowledge, we are the first to combine adaptive threshold extraction with the threshold-independent form of GP distribution in a neural forecasting model. Additionally, We incorporate an autoencoder and a quantile loss function, resulting in further improvements in the accuracy of existing approaches.
\section{PRELIMINARIES}
\subsection{Problem Statement}
\label{Problem Statement}
Given a dataset of $N$ sequences, each containing $n$ predictors within a time window of length $T$, ie., $(\mathbf{X}^{(i)}_{1:T},\mathbf{y}^{(i)}_{T+1}~~ |~~ i \in \{1,\dots,N\})$, where $\mathbf{X}^{(i)}_{1:T}=[\mathbf{x}^{(i)}_{1},\dots,\mathbf{x}^{(i)}_T] \in \mathbb{R}^{n\times T}$ denotes the $i^{th}$ sequence of $n$ predicators for a window size $T$. And we use $\mathbf{x}^{(i)}_{t} = (x_{t,1}^{(i)},x_{t,2}^{(i)},\dots,x_{t,n}^{(i)})^\intercal \in \mathbb{R}^{n}$  to denote a vector of $n$ predictor series input at each time step $t$. $\mathbf{y}^{(i)}_{T+1} \in \mathbb{R}$ denotes the corresponding target values. Our goal  is to develop a model that can accurately forecast the output variable $\mathbf{y}^{(i)}_{T+1}$ given the input time series data $\mathbf{X}^{(i)}_{1:T}$.

\subsection{Generalized Pareto Distribution}
In this work, the GP distribution is used to model rainfall data, supported by literature indicating its effectiveness in capturing rainfall exceedances \cite{coles2003fully}. The GP distribution governs the distribution of exceedances over a predefined threshold $u$. The GP distribution has two parameters, and the CDF of the distribution is given by:
\begin{equation}
\label{tradition_GPD_CDF}
    F_{\xi_u,\sigma_u}(y)=\begin{cases}1-\left(1+\xi_u \frac{y - u}{\sigma_u}\right)^{-1/\xi}&\text{for }\xi_u\neq0,\\ 1-\exp({-\frac{y - u}{\sigma_u})}&\text{for } \xi_u=0.\end{cases}
\end{equation}
where $u$ is the predefined threshold value, $\sigma_u$ is the scale parameter, and $\xi_u$ is the shape parameter.  The subscript $u$ is used to label parameter estimates based on the threshold used. The parameter $\xi_u$ determines the tail behavior of the distribution. For any given $\xi_u$, the scale parameter $\sigma_u$ is responsible for governing the mean of the exceedances beyond the threshold $u$.

\begin{figure*}[h]
  \centering
  \includegraphics[width=\linewidth]{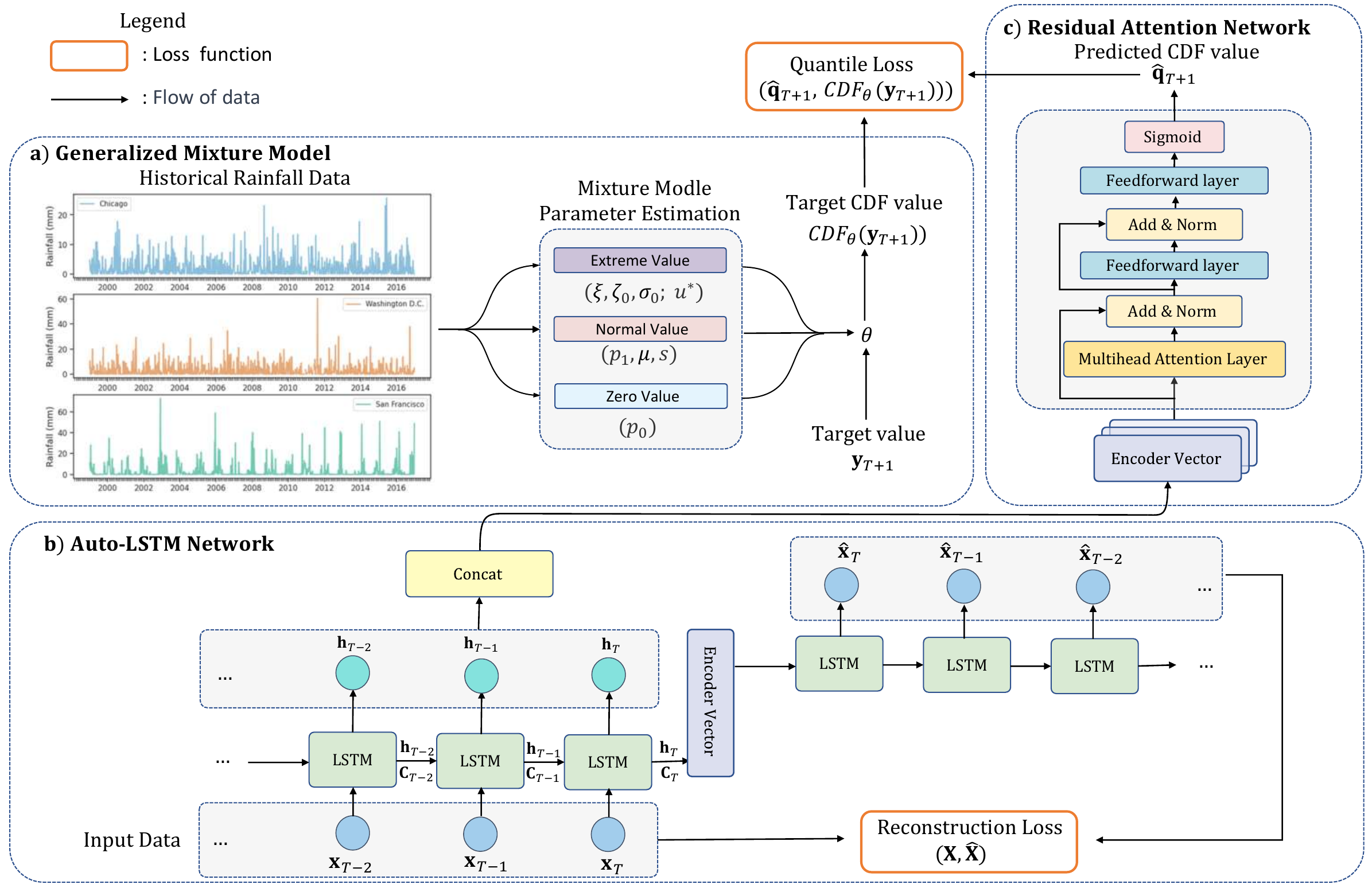}
  \caption{An overview of the architecture of the proposed framework:
\textbf{a)} Generalized Mixture Model: Responsible for modeling the training data. The parameter set $\theta$ of the generalized mixture model is utilized in the subsequent quantile loss function.\quad
\textbf{b)} Auto-LSTM Network: A feature extraction module that  applies an LSTM encoder and decoder. The reconstruction loss is utilized to ensure minimal information loss.\quad
\textbf{c)} Residual Attention Network: A quantile forecaster module comprising temporal attention mechanisms and residual connections. Quantile loss is used during training.}
\label{Model Overview}
\end{figure*}

\section{PROPOSED METHOD}

\subsection{Generalized Mixture Model}
\label{Mixture Model}
This section primarily introduces the proposed generalized mixture model. It combines three probability distributions, each modeling a different range of the target variable. We will discuss the modeling approaches for the extreme values, normal values, and zero values separately. 

\subsubsection{Extreme Value Part}
The traditional GP distribution in Equation \ref{tradition_GPD_CDF} has two parameters ($\xi_u$, $\sigma_u$) with a predefined threshold $u$. The variation of $\sigma_u$ with $u$ can lead to unstable fits if $u$ is not chosen wisely. Ideally, $u$ should be large enough for a reliable GP approximation yet small enough for low estimation variance.

\begin{figure}[t]
  \centering
  \includegraphics[width=\linewidth]{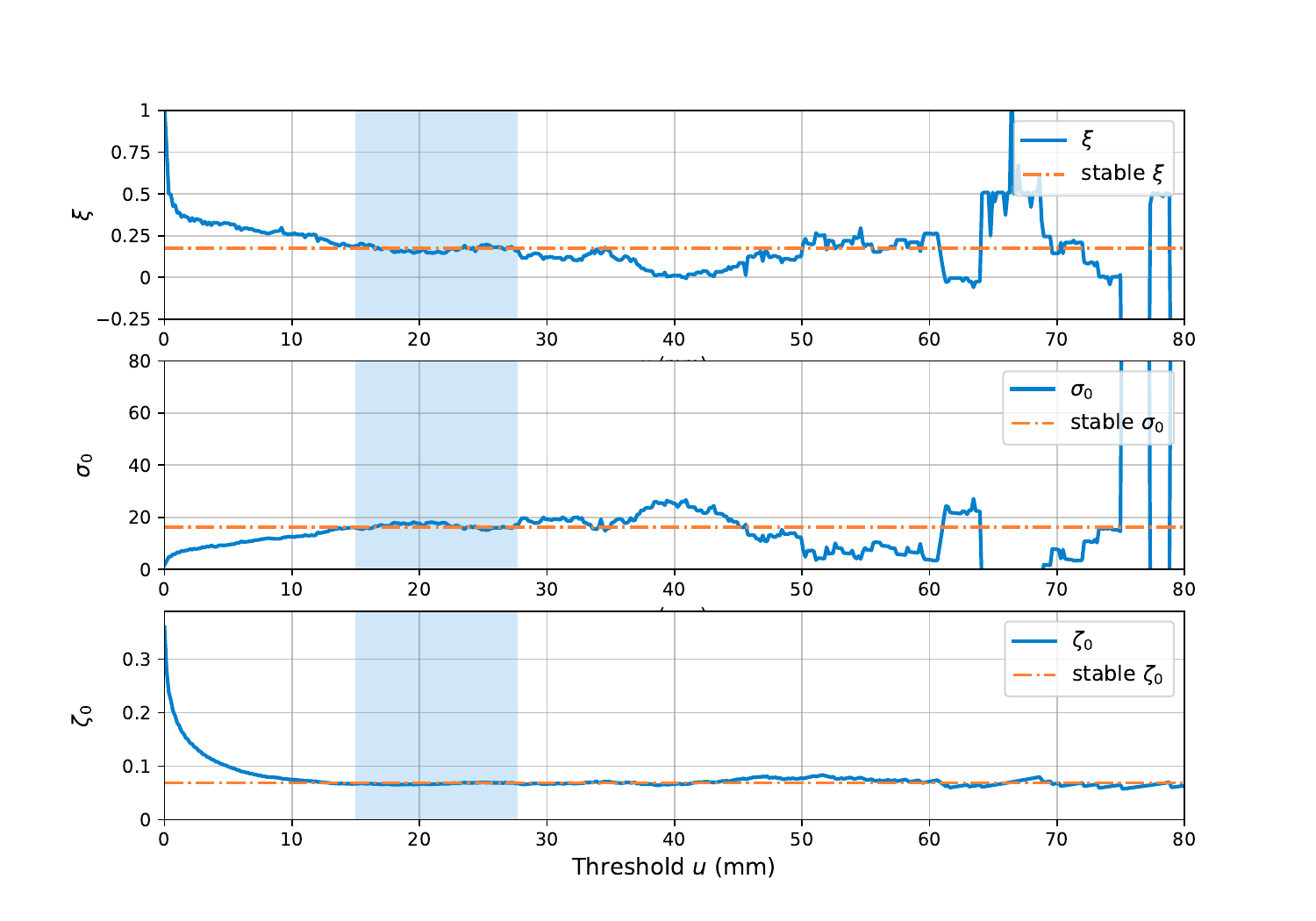}
  \caption{Example of applying the threshold-independent GP distribution form to estimate the parameters for real-world precipitation. The estimates stabilize between 15 mm and 28 mm (blue region). The median values within this region are the estimated parameters (dashed line), suggesting $u^* \approx 15$ mm as the optimum threshold.}
  \label{parameters_}
\end{figure}

To circumvent the challenges of threshold selection, we adopted a threshold-independent distribution form, with its CDF defined as:
\begin{equation}
\label{CDF_u_invarient}
    F_{\xi,\sigma_0, \zeta_0}(y)=\left\{\begin{array}{lr}1-\zeta_0\left(1+\xi\frac{y}{\sigma_0}\right)^{-1/\xi}&\text{for }\xi\neq0\\ 1-\zeta_0\exp\left(-\frac{y}{\sigma_0}\right)&\text{for }\xi=0\end{array}\right.
\end{equation}
, where 
\begin{equation}
    \zeta_0=
\begin{cases}
\zeta_u\left(1+\xi_u\frac{u}{\sigma_0}\right)^{1/\xi}=\zeta_u\left(1-\xi_u\frac{u}{\sigma_u}\right)^{-1/\xi}&\text{for }\xi_u\neq0\\ 
\zeta_u\exp(\frac{u}{\sigma_0})=\zeta_u\exp(\frac{u}{\sigma_u})&\text{for }\xi_u=0
\end{cases}
\end{equation}
and 
\begin{equation}
\sigma_0=\sigma_u-\xi_u u,\quad \forall \xi_u
\end{equation}
$(\xi,\sigma_0, \zeta_0 )$ are the parameters of the distribution in the reparameterized form that are theoretically invariant to changes in the threshold value, as depicted in Figure \ref{parameters_}. Since the GP distribution is employed to describe the extreme values, it can provide the best fit to the data when $u$ surpasses a specific optimal threshold $u^*$.  

To identify the optimal threshold $u^*$, we estimated parameters $(\xi,\sigma_0, \zeta_0 )$ across increasing thresholds, as Figure \ref{parameters_} shows. The estimates stabilize between 15 mm and 28 mm, and the left edge of the stabilized region is chosen as the optimal threshold. Notably, an excessively large threshold causes the estimates to begin fluctuating visibly and exhibit increased variability, highlighting the method's ability to secure stable estimates with ample samples.

\subsubsection{Non-extreme Value Part}
Due to the zero-inflated phenomenon in the rainfall data \cite{beyond_point_prediction}, we use the hurdle model to model zeros and non-zeros for values below the threshold $u^*$. Hurdle models \cite{mullahy1986specification} can be viewed as a two-component mixture model consisting of a zero mass and the positive observations component following a truncated count distribution, such as truncated Poisson or truncated log-normal distribution. The general structure of a hurdle model is given by
\begin{equation}
    P\left(Y=y\right)=\left\{\begin{array}{ll}p_0 & y=0, \\ \left(1-p_0\right) \frac{f\left(y; \Phi \right)}{1-f\left(y=0 ; \Phi \right)} & y>0\end{array}\right.
\end{equation}
, where $p_0$ is the probability of a subject belonging to the zero component. And $f(y; \Phi)$ represents the Probability Density Function (PDF) of a regular count distribution with parameters $\Phi$. We found that fitting the non-zero and non-extreme values with a log-normal distribution yielded the best results for the rainfall dataset.

All three distribution components have been introduced: the Hurdle model, the log-normal distribution, and the GP distribution. By integrating these components, we can establish a comprehensive CDF function for the mixture distribution. The following equation provides the complete CDF function of the generalized mixture model:
\begin{equation}
CDF_{\theta}(y) =
\mathbb{I}_{(0,u^*)}[p_0+p_1\cdot\frac{F_{L}(y;\mu,s)}{{F_{L}(u^*;\mu,s)}}] + \mathbb{I}_{[u^*,+\infty)}  F_{\xi,\sigma_0, \zeta_0}(y)
\end{equation}
, where $\mathbb {I}_{(\cdot)} $is an indicator function. $F_{L}(y;\mu,s) = \frac{1}{2}\left[1+\operatorname{erf}\left(\frac{\ln y-\mu}{s\sqrt{2}}\right)\right]$ is the CDF of log-normal distribution, $\operatorname{erf}(x)$ is the  error function defined as $\operatorname{erf}(x) \equiv \frac{2}{\sqrt{\pi}} \int_{0}^{x} e^{-t^{2}} d t$. $p_0$ and $p_1$ are the probabilities of zero rainfall and non-zero non-extreme values, respectively. The reason for dividing by ${F_{L}(u^*;\mu,s)}$ in the second term of the CDF is to scale it, ensuring the continuity of the CDF function at the breakpoints. The parameters set $\theta$ (including the automatically extracted optimal threshold $u^*$) are given as follows:
$$
\theta = (p_0;~ p_1,\mu,s;~ \xi,\zeta_0,\sigma_0;~ u^*) 
$$
We use these parameters to construct the CDF of the train data and apply it to the rainfall data. Our model is built on this novel CDF formulation to capture the overall distribution. The schematic for parameter extraction of the distribution is depicted in part \textbf{a)} of Figure \ref{Model Overview}.

\subsection{Model Architecture}
Inspired by \cite{hubner2010dual}, we proposed a dual-stage attention-based model. Parts \textbf{b)} and \textbf{c)} of Figure \ref{Model Overview} present the schematic illustration of the proposed model architecture, comprising Auto-LSTM Network, a gross feature extraction module, and residual Attention Network, a fine-grained extraction module. We will provide a detailed explanation of the network architecture below.

\subsubsection{Auto-LSTM Network: Feature Extraction Modules}

Our proposed model starts with an LSTM-based Autoencoder (Auto-LSTM) network, tailored to capture the temporal dependencies of input sequences and extract major features. The model has both an encoder and a decoder made of LSTM layers. The LSTM cell is to learn a mapping from the input $\mathbf{x}_t$ to the hidden state $\mathbf{h}_t$ with $\mathbf{h}_{t}=f\left(\mathbf{h}_{t-1}, \mathbf{x}_{t}\right)
$, where $\mathbf{h}_{t} \in \mathbb{R}^{m}$ is the hidden state of the LSTM cell at time $t$, and $m$ is the size of the hidden state. The final hidden state $\mathbf{h}_T$ of the encoder is used as the initial state for the decoder, as it encapsulates the essential information of the time series. The decoder is trained to reconstruct the time series in reverse order (similar to \cite{sutskever2014sequence} ). 

The encoder defines a feature-learning function $\mathbf{h} = f_{\theta_e} (\mathbf{X})$, and the decoder is trained to reconstruct the original input from the encoded vector, represented by $\widehat{\mathbf{X}} = g_{\theta_d}(\mathbf{h})$.   $\theta_e$ and $\theta_d$ are the parameter sets of the encoder and decoder respectively. The objective of the Auto-LSTM architecture is to minimize the reconstruction loss function, which could be written as
\begin{equation}
\mathit{Reconstruction~Loss} \triangleq \frac{1}{N}\sum_{i=1}^{N}\left\|\mathbf{X}^{(i)}-\widehat{\mathbf{X}}^{(i)}\right\|^2
\end{equation}
, where $\mathbf{X}^{(i)} = g_{\theta_d}\left(f_{\theta_e}\left(\mathbf{X}^{(i)}\right)\right)$. By incorporating the Auto-LSTM network as the initial module of our proposed model, we effectively minimize information loss while extracting fundamental features for subsequent prediction tasks.

\subsubsection{Residual Attention Network: Forecaster Modules}
The attention mechanism has been effectively incorporated into numerous fields \cite{mirsamadi2017automatic, Vaswani,xie2019speech}. The main idea of the attention mechanism is to focus more significantly on a certain distinction in weights. To predict the target value $\textbf{y}_{T+1}$, we calculate the weighted attention of the output of LSTM. Inspired by Vaswani \cite{Vaswani}, we apply a multi-head attention scheme, which helps linearly project the LSTM output into different subspaces with reduced dimensions.

Specifically, we utilize the last LSTM output $\textbf{h}_T$ for Queries to enable a global view, and all outputs $\textbf{h}_{1:T}$ for Keys and Values to capture the specific details at each timestep of the sequence. The attention mechanism takes a set of vectors as input: 
\begin{equation}
\begin{aligned}
    \textbf{Q}_i &= \textbf{Q} \cdot \textbf{W}_i^Q = \textbf{h}_T    \cdot \textbf{W}_i^Q \in \mathbb{R}^{1\times \frac{m}{d}}\\
    \textbf{K}_i &= \textbf{K} \cdot \textbf{W}_i^K = \textbf{h}_{1:T}\cdot \textbf{W}_i^K \in \mathbb{R}^{T\times \frac{m}{d}}\\
    \textbf{V}_i &= \textbf{V} \cdot \textbf{W}_i^V = \textbf{h}_{1:T}\cdot \textbf{W}_i^V \in \mathbb{R}^{T\times \frac{m}{d}}
\end{aligned}
\end{equation}
, where  $\textbf{Q} = \textbf{h}_T \in \mathbb{R}^{1 \times m} ,\textbf{K} = \textbf{h}_{1:T}\in \mathbb{R}^{T \times m} ,\textbf{V} = \textbf{h}_{1:T} \in \mathbb{R}^{T \times m}$ is Queries, Keys and Values separately. $d$ is the number of attention heads. $\textbf{W}_i^Q,\textbf{W}_i^K,\textbf{W}_i^V \in \mathbb{R}^{m\times \frac{m}{d}}$ are the learnable projection matrices for the $i$-th head. 

Mathematically, the $i$-th attention head, computed using the scaled dot product attention mechanism, is given by:
\begin{equation}
\label{attention_qkv}
    \text{head}_{i}= \operatorname{Attention}(\textbf{Q}_i,\textbf{K}_i,\textbf{V}_i)=\operatorname{Softmax}\left(\textbf{Q}_i \textbf{K}_i^{\top} / \sqrt{\frac{m}{d}}\right) \textbf{V}_i
\end{equation} 
, where $\text{head}_{i} \in  \mathbb{R}^{1\times \frac{m}{d}}$. This mechanism enables selective attention to different sequence parts based on the similarity between the queries and keys. 

The multi-head attention, which aggregates these heads by projecting queries, keys, and values into different sub-spaces, is then:
\begin{equation}
\operatorname{Multihead}(\textbf{Q}, \textbf{K}, \textbf{V})=\operatorname{Concat}\left(\operatorname{head}_{1}, \ldots, \operatorname{head}_{d}\right) \textbf{W}^{O}
\end{equation}
, where $W^{O} \in \mathbb{R}^{m \times m}$ is a learnable matrix for combining attention head outputs.

As depicted in part \textbf{b)} of Figure \ref{Model Overview}, the multihead attention output undergoes a two-stage pipeline. Each stage includes an Add \& Norm layer for addressing the vanishing gradient issue and boosting training speed, and a Feedforward layer for refining features. After two repetitions, the result is an intermediate output. Finally, a sigmoid function ensures the predicted output $\hat{\textbf{q}}_{T+1}$ is between $[0, 1]$.

\subsection{Objective Function}
To train the forecaster modules incorporating distributional information, we project both predicted and target values onto the CDF subspace. Instead of predicting $\textbf{y}_{T+1}$ directly, we aim for its quantile, $\textbf{q}_{T+1} = CDF_{\theta}(\textbf{y}_{T+1})$. It is motivated by the fact that this transformation reduces the large variations in the actual time series. Consequently, it results in an output with less volatility and provides a more evenly distributed focus on extreme events during the prediction process. The original value, $\textbf{y}_{T+1}$, is retrievable using the quantile function, or $CDF^{-1}$.

With $\hat{\textbf{q}}_{T+1}\in[0,1]$ as the forecaster's output, we minimize the quantile loss between $\hat{\textbf{q}}_{T+1}$ and $\textbf{q}_{T+1}$, defined as:
\begin{equation}
\begin{split}
\mathit{Quantile~Loss}_\tau \triangleq \frac{1}{N}\sum_i^N\max  (\tau(\textbf{q}_{T+1}^{(i)} -\hat{\textbf{q}}_{T+1}^{(i)}),& \\
 (\tau-1)(\textbf{q}_{T+1}^{(i)} -\hat{\textbf{q}}_{T+1}^{(i)}))&
\end{split}
\end{equation}

, where $\tau$ denotes the quantile level. During training, both the feature extraction and forecaster modules are trained concurrently, optimizing the weighted sum of both losses. The comprehensive loss function is:
\begin{equation}
\mathcal{L} = w\cdot\mathit{Reconstruction~Loss} + (1-w)\cdot \mathit{Quantile~Loss}_\tau
\end{equation}
, where $w$ is a hyperparameter representing the trade-off between minimizing reconstruction and quantile loss.

\section{Experimental Evaluation}
\subsection{Dataset}
We evaluate our model's performance using actual precipitation measurements from major cities across the United States. The predictor variables consist of precipitation forecasts generated by an 11-member ensemble from the SubX project,  which provides a publicly available database of 21 years of historical forecasts (1999–2020). All forecasts include daily values for at least 30 days beyond the initialization date. We compute the rolling 3-day average of each ensemble member for the first nine days, resulting in a prediction time window of 7 days. And our target variable is observed precipitation obtained from NLDAS-2, specifically the average observed precipitation 10-12 days in advance. 
\subsection{Baselines}
We compare the performance of our proposed model with several baseline methods. Among them, the first three models are dedicated to extreme value prediction, while the remaining three are conventional time series forecasting models.
\begin{enumerate}[leftmargin=*]

    \item \textbf{Wilson} \cite{beyond_point_prediction}: A deep learning-based framework for forecasting extreme values that unifies a mixture of distributions for different event types and allows for dynamic threshold setting without retraining. 
    \item \textbf{Ding} \cite{Ding}:  A baseline model for time series prediction that incorporates extreme values. Ding utilizes an EVT-motivated loss function and a memory module to capture and model extreme events in the time series.
    \item \textbf{Vandal} \cite{vandal}: A baseline model that utilizes a discrete-continuous Bayesian Deep Learning (BDL) approach for uncertainty quantification and distribution prediction. 
    \item \textbf{LSTM} \cite{LSTM}: Long Short-Term Memory, a popular RNN baseline used in time series prediction tasks.
    \item \textbf{TST} \cite{TST}: Time Series Transformer, a Transformer-based framework for time series prediction. 
    \item \textbf{FCN} \cite{FCN}: Fully Convolutional Network, which utilizes a fully convolutional architecture to extract temporal features.
 
\end{enumerate}
\begin{figure}[t]
  \centering
  \includegraphics[width=\linewidth]{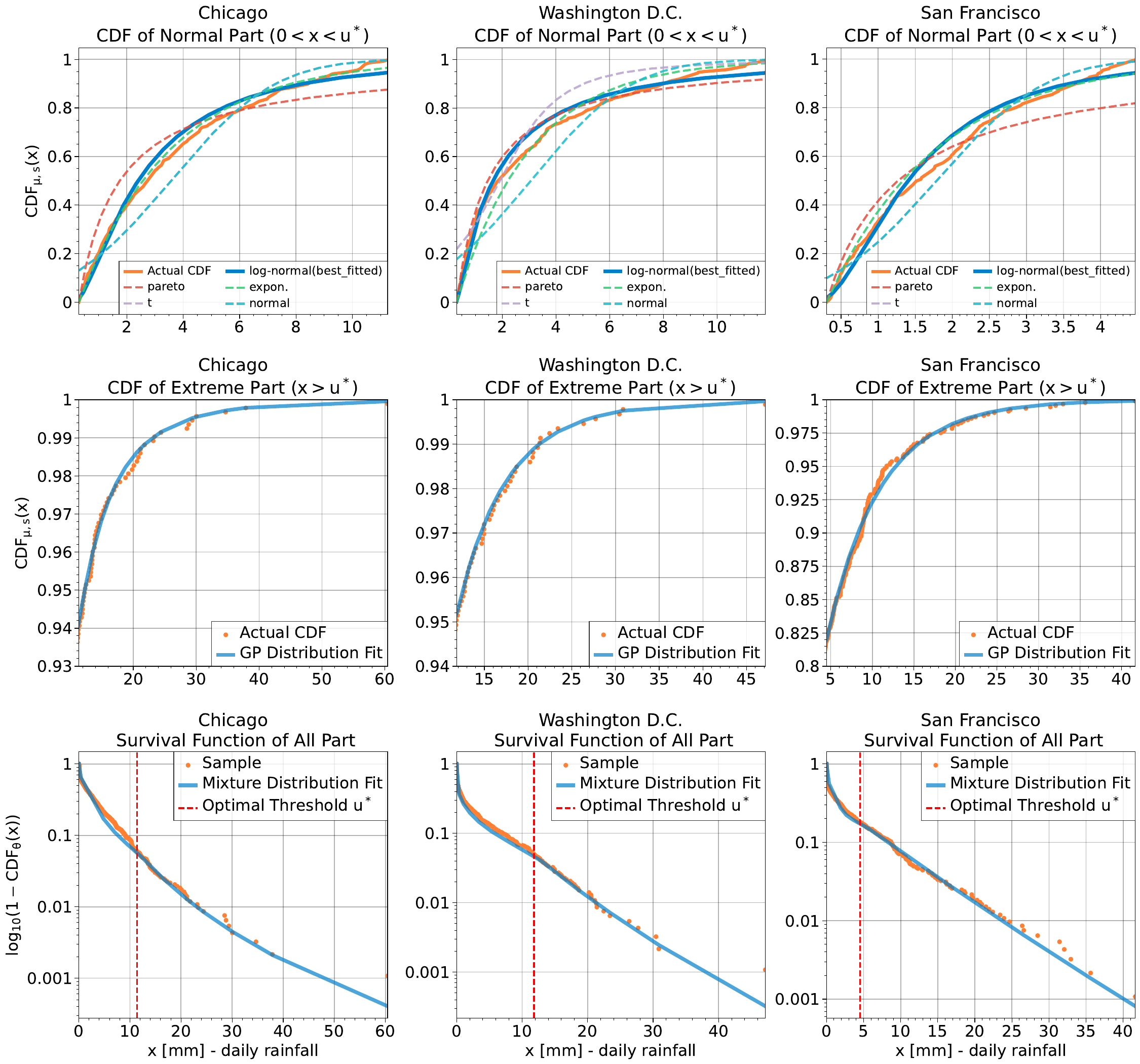}
  \caption{The fitting result of three cities.}
\label{mix_image}
\end{figure}

\begin{table*}[htbp]
\caption{Results of proposed DEMMA and baseline models}
\resizebox{\textwidth}{!}{
\label{table_results}
\begin{tabular}{ccccccccc}
\hline
                                                                            &                & \textbf{DEMMA}                               & \textbf{Wilson}\cite{beyond_point_prediction}                             & \textbf{Ding}\cite{Ding}                               & \textbf{Vandal}\cite{vandal}                             & \textbf{LSTM}\cite{LSTM}                              & \textbf{TST}\cite{TST}                     & \textbf{FCN}\cite{FCN}                        \\ \hline
                                                                            & Extreme\_RMSE    & {\color[HTML]{FFC000} \textbf{4.449±0.879}}  & {\color[HTML]{CD7F32} 4.706±0.842}          & 4.798±0.944                                 & {\color[HTML]{A9A9A9} 4.657±0.907}          & 5.216±0.741                                 & 5.035±0.701                         & 5.180±0.746                         \\
                                                                            & Moderate\_RMSE & {\color[HTML]{FFC000} \textbf{3.521±0.662}}  & {\color[HTML]{A9A9A9} 3.575±0.662}          & {\color[HTML]{CD7F32} 3.620±0.690}          & 3.652±0.702                                 & 3.836±0.570                                 & 3.864±0.483                         & 3.888±0.536                         \\
                                                                            & Zero\_RMSE     & 2.401±0.471                                  & 1.934±0.189                                 & 1.886±0.133                                 & 2.252±0.183                                 & {\color[HTML]{FFC000} \textbf{1.050±0.132}} & {\color[HTML]{CD7F32} 1.614±0.216}  & {\color[HTML]{A9A9A9} 1.512±0.289}  \\
\multirow{-4}{*}{Chicago}                                                   & Total\_RMSE    & {\color[HTML]{FFC000} \textbf{3.208±0.545}}  & {\color[HTML]{A9A9A9} 3.261±0.558}          & {\color[HTML]{CD7F32} 3.294±0.550}          & 3.373±0.574                                 & 3.388±0.480                                 & 3.464±0.402                         & 3.478±0.443                         \\ \hline
                                                                            & Extreme\_RMSE    & {\color[HTML]{FFC000} \textbf{7.658±1.954}}  & 8.023±1.737          & {\color[HTML]{A9A9A9} 7.936±2.050}          & {\color[HTML]{CD7F32} 7.977±2.121}          & 8.699±1.801                                 & 8.461±2.193                         & 8.405±2.116                         \\
                                                                            & Moderate\_RMSE & {\color[HTML]{CD7F32} 6.099±1.370}           & {6.127±1.194}          & {\color[HTML]{FFC000} \textbf{6.070±1.368}} & {\color[HTML]{A9A9A9} 6.093±1.404} & 6.542±1.208                                 & 6.486±1.475                         & 6.450±1.398                         \\
                                                                            & Zero\_RMSE     & 3.464±0.392                                  & 2.590±0.834                                 & {2.584±0.581}          & {\color[HTML]{CD7F32} 2.502±0.440}          & {\color[HTML]{FFC000} \textbf{1.439±0.089}} & 2.898±0.826                         & {\color[HTML]{A9A9A9} 2.458±0.616}  \\
\multirow{-4}{*}{\begin{tabular}[c]{@{}c@{}}Washington\\ D.C.\end{tabular}} & Total\_RMSE    & {\color[HTML]{A9A9A9} 5.370±1.114}           & { 5.394±1.063}          & {\color[HTML]{CD7F32} 5.384±1.206}          & {\color[HTML]{FFC000} \textbf{5.368±1.191}} & 5.617±1.122                                 & 5.779±1.221                         & 5.657±1.227                         \\ \hline
                                                                            & Extreme\_RMSE    & {\color[HTML]{FFC000} \textbf{10.121±3.331}} & 10.815±3.762                                & 11.022±4.163                                & 11.455±4.201                                & {\color[HTML]{CD7F32} 10.416±4.285}         & 12.191±4.472                        & {\color[HTML]{A9A9A9} 10.350±3.712} \\
                                                                            & Moderate\_RMSE & {\color[HTML]{FFC000} \textbf{9.971±2.941}}  & {\color[HTML]{A9A9A9} 9.989±3.269}          & 10.424±3.603                                & 11.397±3.770                                & {\color[HTML]{CD7F32} 10.160±3.854}         & 11.593±4.103                        & 10.194±3.271                        \\
                                                                            & Zero\_RMSE     & 2.429±1.995                                  & 1.870±1.189                                 & {\color[HTML]{CD7F32} 1.275±0.395}          & {\color[HTML]{FFC000} \textbf{0.912±0.175}} & {\color[HTML]{A9A9A9} 1.000±0.286}          & 1.758±1.702                         & 1.601±0.577                         \\
\multirow{-4}{*}{\begin{tabular}[c]{@{}c@{}}San \\ Francisco\end{tabular}}  & Total\_RMSE    & {\color[HTML]{A9A9A9} 6.613±1.811}           & 6.961±2.430                                 & 6.937±2.641                                 & 7.198±2.625                                 & {\color[HTML]{FFC000} \textbf{6.558±2.666}} & 7.770±2.558                         & {\color[HTML]{CD7F32} 6.623±2.271}  \\ \hline
                                                                 
                                                                            & Extreme\_RMSE    & {\color[HTML]{FFC000} \textbf{9.926±2.823}}  & {\color[HTML]{CD7F32} 10.326±2.732}         & {\color[HTML]{000000} 10.440±2.503}         & {\color[HTML]{A9A9A9} 10.091±2.716}         & {\color[HTML]{000000} 11.040±2.491}         & {\color[HTML]{000000} 10.777±2.785} & {\color[HTML]{000000} 10.632±2.743} \\
                                                                            & Moderate\_RMSE & {\color[HTML]{FFC000} \textbf{7.782±2.197}}  & {\color[HTML]{CD7F32} 8.018±2.167}          & {\color[HTML]{000000} 8.068±2.014}          & {\color[HTML]{A9A9A9} 7.873±2.113}          & {\color[HTML]{000000} 8.498±2.032}          & {\color[HTML]{000000} 8.504±2.163}  & {\color[HTML]{000000} 8.260±2.173}  \\
                                                                            & Zero\_RMSE     & {\color[HTML]{000000} 2.927±0.353}           & {\color[HTML]{000000} 2.247±0.648}          & {\color[HTML]{000000} 2.110±0.386}          & {\color[HTML]{000000} 2.801±0.710}          & {\color[HTML]{FFC000} \textbf{1.359±0.241}} & {\color[HTML]{CD7F32} 1.941±0.393}  & {\color[HTML]{A9A9A9} 1.522±0.301}  \\
\multirow{-4}{*}{Atlanta}                                                   & Total\_RMSE    & {\color[HTML]{FFC000} \textbf{6.656±1.669}}  & {\color[HTML]{CD7F32} 6.768±1.649}          & {\color[HTML]{000000} 6.791±1.560}          & {\color[HTML]{A9A9A9} 6.730±1.592}          & {\color[HTML]{000000} 7.082±1.583}          & {\color[HTML]{000000} 7.135±1.656}  & {\color[HTML]{000000} 6.897±1.705}  \\ \hline
                                                                            & Extreme\_RMSE    & {\color[HTML]{FFC000} \textbf{8.241±0.516}}  & 12.045±1.999                                & 14.188±2.115                                & 14.307±1.329                                & {\color[HTML]{A9A9A9} 10.849±0.789}         & 12.672±0.893                        & {\color[HTML]{CD7F32} 11.153±1.078} \\
                                                                            & Moderate\_RMSE & {\color[HTML]{FFC000} \textbf{7.661±0.283}}  & 8.670±1.113                                 & 9.923±1.210                                 & 10.022±0.971                                & {\color[HTML]{A9A9A9} 7.845±0.581}          & 10.439±0.416                        & {\color[HTML]{CD7F32} 8.509±0.604}  \\
                                                                            & Zero\_RMSE     & 5.463±1.164                                  & 3.026±2.236                                 & {\color[HTML]{FFC000} \textbf{1.916±2.946}} & {\color[HTML]{A9A9A9} 2.114±0.503}          & {\color[HTML]{CD7F32} 2.511±0.746}          & 4.813±3.544                         & 2.873±0.648                         \\
\multirow{-4}{*}{Seattle}                                                   & Total\_RMSE    & {\color[HTML]{A9A9A9} 7.365±0.336}           & 8.061±1.016                                 & 9.130±1.036                                 & 9.231±0.848                                 & {\color[HTML]{FFC000} \textbf{7.267±0.480}} & 9.806±0.556                         & {\color[HTML]{CD7F32} 7.891±0.494}  \\ \hline
                                                                            & Extreme\_RMSE    & {\color[HTML]{FFC000} \textbf{7.547±3.347}}  & {\color[HTML]{CD7F32} 8.495±3.367}          & 9.013±3.820                                 & 9.433±3.106                                 & {\color[HTML]{A9A9A9} 8.365±3.601}          & 8.861±3.295                         & 8.580±2.960                         \\
                                                                            & Moderate\_RMSE & {\color[HTML]{FFC000} \textbf{6.579±2.257}}  & {\color[HTML]{CD7F32} 6.898±2.634}          & 7.244±2.844                                 & 7.689±2.419                                 & {\color[HTML]{A9A9A9} 6.809±2.723}          & 7.570±2.291                         & 7.039±2.184                         \\
                                                                            & Zero\_RMSE     & 2.318±0.862                                  & 2.189±0.672                                 & 1.604±0.718                                 & {\color[HTML]{A9A9A9} 1.318±1.092}          & {\color[HTML]{FFC000} \textbf{1.203±0.388}} & 1.704±0.872                         & {\color[HTML]{CD7F32} 1.412±0.613}  \\
\multirow{-4}{*}{Miami}                                                     & Total\_RMSE    & {\color[HTML]{FFC000} \textbf{5.425±1.693}}  & {\color[HTML]{CD7F32} 5.647±2.105}          & 5.848±2.223                                 & 6.161±2.001                                 & {\color[HTML]{A9A9A9} 5.459±2.177}          & 6.124±1.798                         & 5.674±1.741                         \\ \hline
                                                                            & Extreme\_RMSE    & {\color[HTML]{FFC000} \textbf{4.591±2.119}}  & {\color[HTML]{A9A9A9} 4.902±2.050}          & 5.115±2.216                                 & {\color[HTML]{CD7F32} 4.904±1.987}          & 5.488±1.988                                 & 5.364±1.995                         & 5.470±1.909                         \\
                                                                            & Moderate\_RMSE & {\color[HTML]{FFC000} \textbf{3.816±1.552}}  & {\color[HTML]{CD7F32} 3.865±1.667}          & 3.943±1.740                                 & {\color[HTML]{A9A9A9} 3.861±1.635}          & 4.143±1.636                                 & 4.238±1.589                         & 4.236±1.573                         \\
                                                                            & Zero\_RMSE     & 2.161±0.331                                  & 2.178±0.374                                 & {\color[HTML]{CD7F32} 2.032±0.470}          & 2.346±0.422                                 & {\color[HTML]{FFC000} \textbf{1.405±0.145}} & 2.573±0.227                         & {\color[HTML]{A9A9A9} 1.864±0.481}  \\
\multirow{-4}{*}{St. Louis}                                                 & Total\_RMSE    & {\color[HTML]{FFC000} \textbf{3.454±1.082}}  & {\color[HTML]{CD7F32} 3.537±1.270}          & 3.543±1.266                                 & {\color[HTML]{A9A9A9} 3.524±1.248}          & 3.611±1.236                                 & 3.871±1.193                         & 3.748±1.206                         \\ \hline
\end{tabular}}
\end{table*}
\subsection{Experimental Setup}
For evaluation purposes, we split the data into separate sets for training, validation, and testing in a ratio of 7:2:1 for all models. The data is standardized to have a mean of zero and a variance of one before being fed into the network. Hyperparameters were selected using grid search. All models were trained for 50 epochs with checkpoints saved at the lowest validation loss. For each dataset, we employed five different random train-validation-test splits and computed the mean and standard deviation for each metric. 
We consider the following evaluation metrics in our experiments. 
\begin{enumerate}[leftmargin=*]
 \item Total\_RMSE:  Rooted mean squared error between the actual and predicted value of each model, providing an assessment of the overall accuracy.
 \item Extreme\_RMSE:  measures the accuracy of predictions for the extreme component, where the target values exceed the 0.6 quantile threshold.
 \item Moderate\_RMSE: measures the accuracy of predictions for the moderate component, where the target values fall within the range of 0 and the 0.6 quantile threshold.
 \item Zero\_RMSE: measures the accuracy of predictions for the zero component, corresponding to target values equal to zero.
\end{enumerate}

\subsection{Experimental Results}
\subsubsection{Fitting Results}

We applied the parameter estimation method introduced in Section \ref{Mixture Model} to the dataset. We presented the fitting results for three cities datasets, as figure \ref{mix_image} shows. 

These graphs demonstrate that our method achieved good fitting results. The top graphs display the CDF fitted using different distributions on the normal part of values. Among them, the log-normal distribution performs the best. The middle graphs present an enlarged view of the empirical CDF and the mixture distribution in the extreme value region, and once again, we can observe a good fitting. The bottom graphs show a logarithmic empirical survival function compared to the obtained mixture distribution. We can observe how the fitting reliably captures the extreme values, even when some optimal thresholds are very low. The optimal threshold $u^*$ is represented by a vertical dashed red line.

\subsubsection{Performance Against Baselines}
Table \ref{table_results}  compares the DEMMA model with the baseline models mentioned earlier. Results are expressed as $\bar{x} \pm s$ for sample mean and standard deviation, with the top three performances for each metric denoted in {\color[HTML]{FFC000} \textbf{gold}}, {\color[HTML]{A9A9A9} silver} and {\color[HTML]{CD7F32} bronze}. 

To assess the prediction accuracy in different regions, we calculated the RMSE of the target values within each region. The DEMMA model outperforms the baseline models in terms of Extreme\_RMSE in the majority of cities' datasets. This fact demonstrates the effectiveness of our model in predicting extreme values. Furthermore, our model does not sacrifice overall predictive performance for extreme value accuracy. It performs at a competitive level in terms of Total\_RMSE, which demonstrates the effectiveness of incorporating a mixture distribution in modeling the overall data. Moreover, we have observed that the Extreme\_RMSE of models designed for extreme value prediction is relatively smaller across most of the datasets, whereas traditional deep models, especially LSTM and FCN, exhibit smaller errors for zero values. However, when considering overall performance, models designed for extreme value prediction still demonstrate superior performance across the majority of datasets.

\section{Conclusion}

In this paper, we propose DEMMA, a deep learning framework for extreme value prediction using a generalized mixture model that combines a hurdle model and a GP distribution tailored for heavy-tailed data. Using a reparametrized, threshold-independent GP distribution, we adaptively extract its parameters for optimal dataset fit. Regarding the network architecture, we employ Auto-LSTM for robust temporal feature extraction and a residual attention network for precision in forecasting quantile values. Tested across diverse rainfall datasets, DEMMA consistently outperforms both leading extreme value predictors and traditional DNN methods in overall and extreme value accuracy.

\bibliographystyle{IEEEtran}
\bibliography{acmart}

\vspace{12pt}

\end{document}